\documentclass[sigconf]{acmart}
\usepackage{amsmath}  
\usepackage{algorithm}
\usepackage{algpseudocode}
\usepackage{bm}
\usepackage{subfig}
\usepackage{tikz}

\usetikzlibrary{shapes,decorations,arrows,calc,arrows.meta,fit,positioning}

\graphicspath{{images/}}

\makeatletter
\renewcommand{\fnum@figure}{Fig. \thefigure}
\makeatother

\AtBeginDocument{%
  }

\begin{document}

\title{A Human-Centered Approach for\\ Bootstrapping Causal Graph Creation}

\author{Minh Q.~Tram}
\email{minh.tram@mavs.uta.edu}
\affiliation{%
  \institution{The University of Texas at Arlington}
  \city{Arlington}
  \state{TX}
  \country{USA}
  \postcode{76019}
}
\author{Nolan B.~Gutierrez}
\email{nolan.gutierrez@mavs.uta.edu}
\affiliation{%
  \institution{The University of Texas at Arlington}
  \city{Arlington}
  \state{TX}
  \country{USA}
  \postcode{76019}
}
\author{William J.~Beksi}
\email{william.beksi@uta.edu}
\affiliation{%
  \institution{The University of Texas at Arlington}
  \city{Arlington}
  \state{TX}
  \country{USA}
  \postcode{76019}
}

\renewcommand{\shortauthors}{Tram et al.}

\begin{abstract}
Causal inference, a cornerstone in disciplines such as economics, genomics, and
medicine, is increasingly being recognized as fundamental to advancing the
field of robotics. In particular, the ability to reason about cause and effect
from observational data is crucial for robust generalization in robotic
systems. However, the construction of a causal graphical model, a mechanism for
representing causal relations, presents an immense challenge. Currently, a
nuanced grasp of causal inference, coupled with an understanding of causal
relationships, must be manually programmed into a causal graphical model. To
address this difficulty, we present initial results towards a human-centered
augmented reality framework for creating causal graphical models. Concretely,
our system bootstraps the causal discovery process by involving humans in
selecting variables, establishing relationships, performing interventions,
generating counterfactual explanations, and evaluating the resulting causal
graph at every step. We highlight the potential of our framework via a physical
robot manipulator on a pick-and-place task. 
\end{abstract}

\keywords{
Human-Robot Interaction, Augmented Reality, Causal Graphs
}

\maketitle

\section{Introduction}
\label{sec:introduction}
Causality allows for deciphering complex relationships and drawing informed
conclusions. This foundational understanding is beginning to make significant
contributions in the field of robotics, reshaping the path of its
advancement \cite{li2023deep}. However, constructing accurate causal models,
including causal inference and graph representations, is a daunting task. This
stems from the inherent intricacy of the interactions, often compounded by a
myriad of variables, which often remain elusive or hidden in the environment. 

A causal graphical model, aka causal graph, serves as a visual mathematical
representation of the relationships between variables in a system. Delineating
the direction and structure of these relationships allows for identifying
potential causes and effects, and disentangling confounding variables
\cite{scholkopf2021toward}. By employing causal graphs, robots can predict the
outcomes of their actions more accurately, make informed decisions in dynamic
environments, and adapt more fluidly to new situations with a deeper
comprehension of the underlying mechanisms
\cite{xiong2016robot,lee2021causal,cao2021reasoning}. 

Parallel to these advancements, the use of simulations and virtual interfaces
offers a more economical means of gathering diverse and valuable data, which is
especially conducive to improving machine learning algorithms
\cite{mittal2023orbit}. For example, the integration of virtual, augmented, and
mixed reality (VAMR) has shown promising results in fostering a more intuitive
and enhanced interaction between human operators and robots (e.g.,
\cite{puljiz2019general,puljiz2020what,tram2023intuitive}). VAMR interfaces
provide a way for operators to have a more immersive control experience,
bridging the gap between the digital and physical realms. 

\begin{figure}
\centering
\vspace{7mm}
\includegraphics[width=\linewidth]{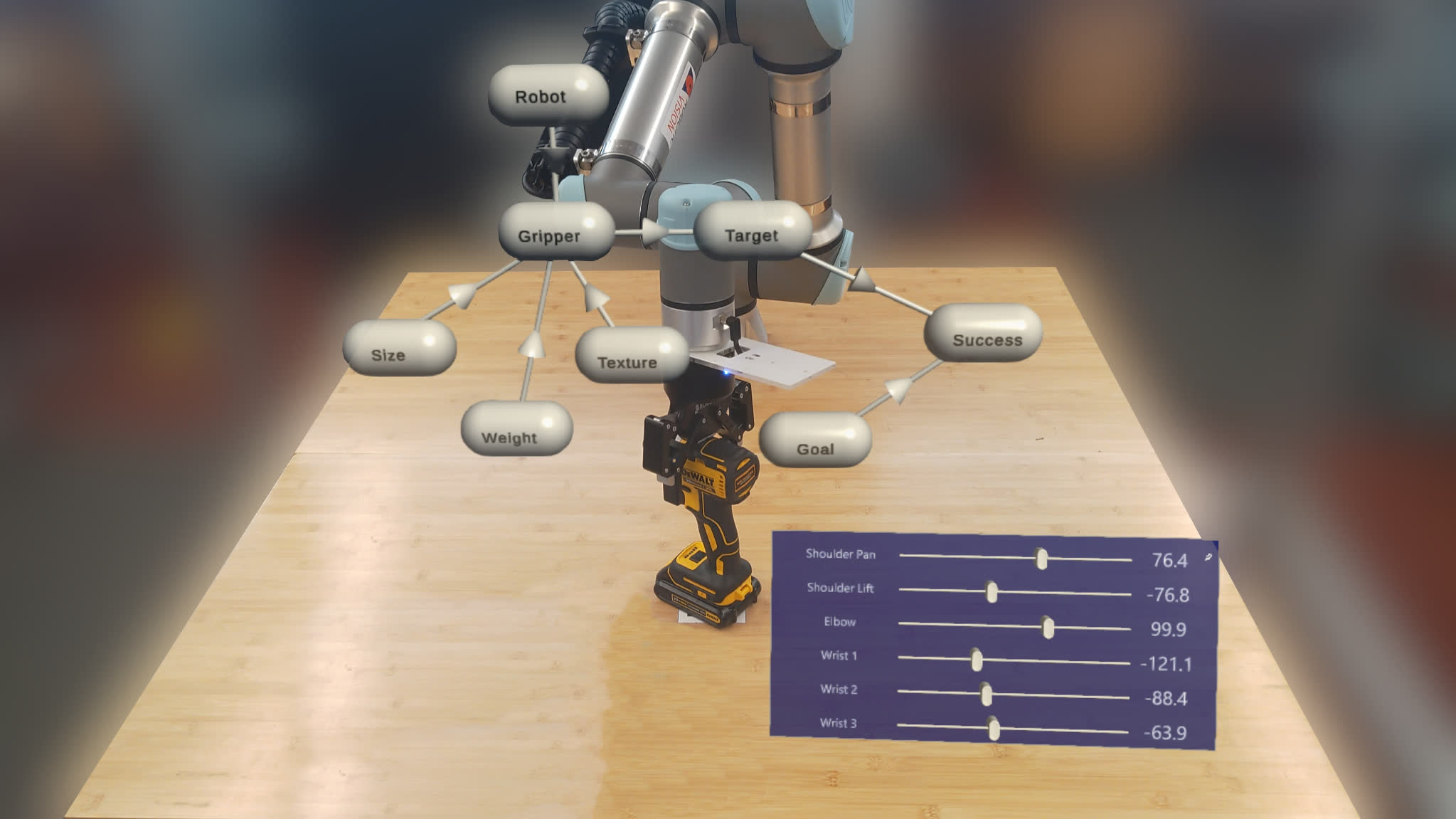}
\caption{The point of view of the operator from the AR overlaid physical
workspace. The operator can interact with the robot and provide additional
context to aid the robot's understanding via on-the-fly construction of causal
graphs.} 
\label{fig:overview}
\end{figure}

In this work we present preliminary results on a new framework that leverages
VAMR technologies to address the challenge of causal graph construction,
Fig.~\ref{fig:overview}. To summarize, our contributions are the following.
\begin{itemize}
  \item We create an augmented reality (AR) interface that allows an operator
  to naturally construct a causal graph on the fly. 
  \item We provide a human-centered approach to highlight critical information
  when establishing a causal graph of a scene.
  \item Our system ensures that a robot prioritizes human insights during the
  initial stages of scene comprehension. 
\end{itemize}

\section{Background and Motivation}
\label{sec:background_motivation}
\subsection{Causal Graphs and Causal Inference}
\begin{figure}
\centering
\begin{tikzpicture}[->,>=stealth',auto,node distance=1cm,
  thick,state/.style={ellipse,draw,font=\LARGE,minimum size = 1.3cm}]
\node [state] (V) {V};
\node [state, left = 1.5cm of V] (B) {B};
\node [state, right = 1.5cm of V] (T) {T};
\path[every node/.style={font=\sffamily\small}]
    (B) edge node [right] {} (V)
    (T) edge[dashed] node [left] {} (V);
\end{tikzpicture}
\caption{A causal graph representing the relationship between a robot's battery
level $B$, the terrain roughness $T$, and the robot's velocity $V$ during
navigation. The solid edge represents a well-established cause and effect,
while the dashed edge represents an indirect or latent confounding variable.}
\label{fig:causal_graph}
\end{figure}
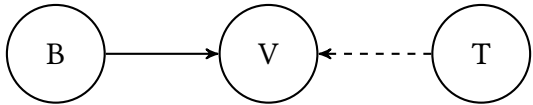

To establish a foundation for causal inference, Pearl introduced a number of
mathematical formulations, notably structural causal models (SCMs),
intervention and do-calculus, and counterfactuals, to rigorously define the
relationships between variables. For example, consider a mobile robot
navigation scenario. The causal graph that represents the relationship between
the robot's velocity, the battery level, and the roughness or difficulty of the
terrain can be represented as a directed acyclic graph (DAG),
Fig.~\ref{fig:causal_graph}. In this scenario, we can represent the
relationships between the variables using an SCM $f$ as
\begin{equation}
  V = f(B,T,U_V),
\end{equation}
where $V$ is the velocity, $B$ is the battery level, $T$ is the terrain
roughness, and $U_V$ is an internal variable that is unobserved, but has an
effect on the velocity of the robot. An intervention in this situation would be
to charge the robot's battery regardless of any other factor hence ensuring it
is always full. This can be represented using do-calculus, i.e., 
\begin{equation}
  P(V\,|\,do(B=\text{full})) = \sum_{T} P(V\,|\,B=\text{full}, T=t)P(T=t).
  \label{eq:do-calculus}
\end{equation}

In \eqref{eq:do-calculus}, estimates of the speed of the robot are given
through the intervention by averaging over all possible values of $T$. Assuming
that there are no confounding relationships between the variables, we have
effectively broken the natural causal relationship between $B$ and $V$, while
leaving the causal relationship between $T$ and $V$ intact. This is where
counterfactuals can help infer the causal relationship between $B$ and $V$
during an intervention, which can aid in reinforcing or disproving causal
relationships between variables. This is important for both pre-planning,
on-the-fly decision making, and insights toward post-incident analysis. By
using this mathematical framework to understand the causal relationships
between variables, we can make informed predictions about a system. For
instance, in the robot navigation scenario we can use these relationships to
pinpoint the root cause of performance issues, pre-charge the robot's battery
if it is expected to go through rough terrain that will drain the battery, or
adjust the robot's behavior to achieve a desired outcome.

\subsection{Motivation}
The use of causal inference is not new in robotics
\cite{brawer2020causal,ho2020actionalperceptual,diehl2023causalbased,ding2023causalaf}.
\textit{Nonetheless, automating the creation of a viable causal graph remains
an underexplored area of research.} Constructing a causal graph is hard due to
the following reasons.
\begin{itemize}
  \item \textbf{Complexity of the environment:} Variables and their causal
  relationships must be identified in dynamic environments while accounting for
  confounding variables.
  \item \textbf{Causation vs. correlation:} The distinction between causation
  and correlation from observational data is not always clear, and definitive
  proof of causation often requires more data.
  \item \textbf{High dimensionality:} The number of variables in an environment
  can be large, which can lead to complex and/or multiple causal graphs.
\end{itemize}

\section{Human-Centered Causal Graphical Models}
\label{sec:human-centered_causal_graphical_models}
In this section, we delineate the specific methodologies and procedures
employed to actualize human-centered causal graphs. First, we present a brief
overview of our framework's components. Then, we provide a detailed description
of each component and the communication pipeline between them.

\subsection{Component Overview and Consideration}
\begin{figure}
\centering
\includegraphics[width=\linewidth]{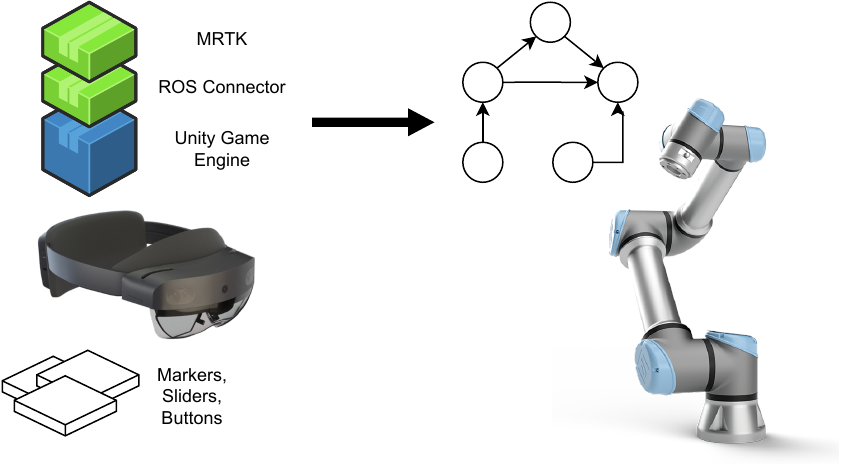}
\caption{The visualization and interaction pipeline of our proposed framework.
The operator can interact with the robot using an overlay interface via various
modes and provide context hinting directly onto the workspace to aid the robot
in its understanding of the scene.}
\label{fig:system_architecture}
\vspace{-4mm}
\end{figure}

Our framework to facilitate robotic understanding through human-centered causal
graph construction is composed of three primary components: (i) an AR-capable
headset, (ii) a software interface to visualize the robot's perception and
allow the operator to interact with the robot, and (iii) a control interface to
realize the operator's commands and interventions. Each of these components
plays a pivotal role in the seamless execution of the system, providing both
unique and complementary functionalities. Fig.~\ref{fig:system_architecture}
details the overall system architecture and the transmission pipeline between
the components.

For the AR headset, we opted for the Microsoft HoloLens 2 due to its
unparalleled capabilities. Its commercial availability ensures ease of access
and its extensive API support, via Microsoft's Mixed Reality Toolkit (MRTK)
\cite{mrtk}, offers a versatile platform for development and integration. The
HoloLens 2 is a self-contained head-mounted display (HMD) that is capable of
spatial mapping and tracking, while also providing a projected holographic
display. It permits the operator to operate in a hands-free, untethered manner.
This feature is crucial for our work for two primary reasons. First, being able
to superimpose additional information directly onto the workspace enhances the
operator's understanding of the robot's perception of the environment. This is
vital when visualizing the implications of the causal graph on the robot's
physical actions in real-time. Second, the ability to operate in a hands-free
manner through different modalities (i.e., direct-hand input, eye-tracking, and
voice commands) allows the operator to swiftly and naturally interact with both
the robot's actions and the underlying causal structure. This ensures that the
operator can manipulate the causal graph, adjust parameters, and interact with
robot perception and environment understanding without the constraints of
traditional input methods.

\subsection{Interaction and Visualization Modalities}
Relying on the HoloLens 2's hand-tracking capabilities, our system allows the
operator to interact with the robot and environment through a number of
mechanisms. This includes direct manipulation of the robot's joint positions
via virtual sliders and command buttons, providing a planning context such as
desired end-effector positions and approach vectors, and the ability to query
forward and inverse kinematics solutions. Furthermore, compared to traditional
VR devices where operator interactions are bound to handheld controllers and
the operator's vision is obstructed by a display, the HoloLens 2 offers greater
interaction and visualization capabilities. 

\subsection{Causal Graph Construction and Intervention}
In our human-centered framework, the focus is primarily on the dynamic
construction and intervention of causal graphs. To this end, we leveraged Unity
\cite{unity} to render causal structures as 3D objects in the AR space allowing
the operator to interact with the causal graph directly. These graphical
constructs, composed of nodes and edges, offer a visual guide that aids the
operator in deciphering the complex interplay between robot actions and
environmental cues. 

Nodes, visualized as tangible entities within the AR landscape, symbolize
specific events or actions that the robot can detect or perform. Their design
incorporates distinct visual features, such as shades, opacities, and
pulsations, communicating their present state or relevance. On the other hand,
edges provide visual insights into the direction, strength, or likelihood of
causal linkages. These are often characterized by variations in their
thickness, texture, or color gradient, allowing users to quickly comprehend
causal dependencies.

Within the immersive world facilitated by the HMD, operators can actively
modify and intervene in the causal relationships by adding, removing, or
modifying the nodes and edges of the current causal structure. The device's
advanced hand-tracking feature supports intuitive gestures, such as pinching,
swiping, or rotating, enabling users to adjust causality strength, node
prominence, or even the direction of a causal link. Furthermore, the mobility
of the HMD lets an operator freely move around the environment and provide
context hints directly onto the robot workspace, accommodating seamless
adjustments and experiments with the causal graph.

\subsection{Physical Robot Actuation and Simulation}
The physical robot actuation and simulation components of our system, driven by
ROS 2 Humble \cite{ros}, facilitate communication between the robot and Unity
software layers. This allows the operator to interact with the robot and
environment through the AR interface. Concretely, it enables the translation of
intricate directives from a causal graph into precise physical movements of the
robot, ensuring adaptability and dexterity in diverse real-world scenarios. 

\section{Evaluation}
\label{sec:evaluation}
\subsection{Causal Graph Creation}
Using a UR5e robot arm provisioned with a Robotiq 2F-85 gripper and tasked with
picking up a finite set of known objects and placing them at a desired
location, we construct a causal graph to represent the relationship between the
robot's actions and the environment. To achieve this, the potential variables
related to our task are represented as follows.
\begin{enumerate}
  \item $Robot$: The robot state.
  \item $Gripper$: The gripper state.
  \item $Target$: The robot's motion planning to achieve the goal position.
  \item $Goal$: The desired object placement location.
  \item $Type$: The object's type.
  \item $Weight$: The object's weight (light - heavy).
  \item $Size$: The object's size (small - large).
  \item $Texture$: The object's texture (smooth - rough).
  \item $Rigidity$: The object's rigidness (soft - hard).
  \item $Success$: The success rate of a pick-and-place sequence.
\end{enumerate}

\begin{figure}
\centering
\begin{tikzpicture}[->,>=stealth',auto,node distance=0.5cm,
  thick,state/.style={ellipse,draw}]
\node [state] (Wg) {Weight};
\node [state, left = of Wg] (Sz) {Size};
\node [state, right = of Wg] (Tx) {Texture}; 
\node [state, right = of Tx] (Rd) {Rigidity}; 
\node [state, above = 1.25cm of $(Tx.west)!0.5!(Wg.east)$] (Ty) {Type};
\node [state, below = 1.25cm of $(Wg.west)!0.5!(Sz.east)$] (Gr) {Gripper};
\node [state, right = 1.25cm of Gr] (Ta) {Target};
\node [state, below = 1cm of Gr] (Ro) {Robot};
\node [state, below= 1cm of Ta] (Go) {Goal};
\node [state, right = 1.7cm of $(Ta.south)!0.5!(Go.north)$] (Sc) {Success};
\path[every node/.style={font=\sffamily\small}]
    (Ty) edge node [below] {} (Sz)
    (Ty) edge node [below] {} (Wg)
    (Ty) edge node [below] {} (Tx)
    (Ty) edge node [below] {} (Rd)
    (Sz) edge node [below] {} (Gr)
    (Wg) edge node [below] {} (Gr)
    (Tx) edge node [below] {} (Gr)
    (Rd) edge node [below] {} (Gr)
    (Gr) edge node [right] {} (Ta)
    (Ro) edge node [above] {} (Gr)
    (Ta) edge node [right] {} (Sc)
    (Go) edge node [right] {} (Sc);
\end{tikzpicture}
\caption{An example of the relationships between the operator-identified
variables for a pick-and-place task. Each edge corresponds to a direct causal
relationship between the connected variables. Changes in the source of an edge
have a direct consequence on the target of an edge.}
\label{fig:pnp_graph}
\vspace{-2mm}
\end{figure}
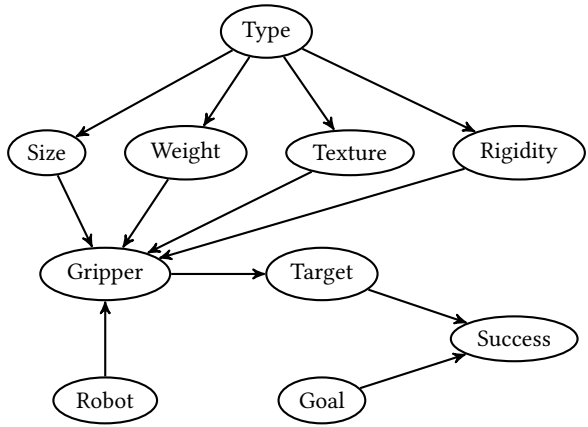

For the purpose of demonstration, we construct this causal graph under the
assumption that external factors (e.g., robot power loss, environmental
disruptions, etc.) do not occur and are insulated from the causal
representation. As shown in Fig.~\ref{fig:pnp_graph}, the relationships between
the variables can be visually represented as a graph for an easier
understanding of potential causality and the interplay between nodes. Within
the graph, each edge represents a direct causal relationship between the robot
and the operating environment. For example, if the weight of the object changes
then it could influence the behavior of the gripper (e.g., the gripper force or
grasp vector).

\begin{figure}
\centering
\includegraphics[width=\linewidth]{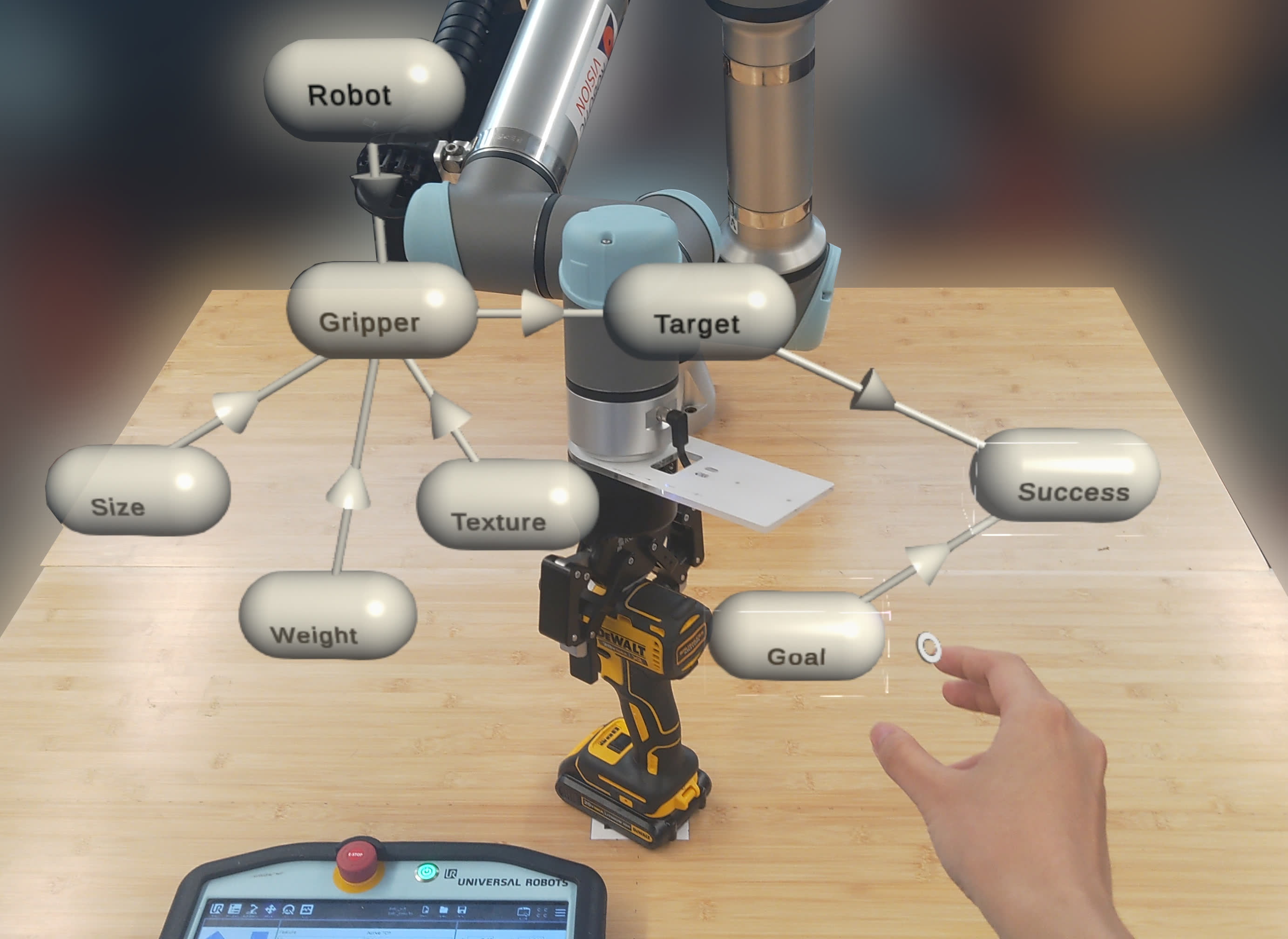}
\caption{The operator interacting with the robot and environment through the AR
interface to construct a causal graph. Nodes and edges are visualized and can
be manipulated directly in the AR space.}
\label{fig:pnp_ar}
\vspace{-2mm}
\end{figure}

Our framework can directly create the causal graph in the AR space,
Fig.~\ref{fig:pnp_ar}. The operator interacts via the AR interface to label the
nodes and edges, set the content of each node, and adjust the strength of each
edge. The operator can also communicate with the robot and the environment
through the AR interface to obtain additional information and context to aid in
the construction of the causal graph. The causal graph can then be parsed and
interpreted by the motion planning backend during the planning step to ensure
that the robot's actions are consistent with the state of the graph.
Algorithm~\ref{alg:graph_consume} demonstrates how the robot consumes a causal
graph generated by the operator during each motion planning step.

\begin{algorithm}
\caption{Human-in-the-loop Causal Graph Iteration}
\label{alg:graph_consume}
\begin{algorithmic}[1] 
\Require Human has initialized the causal graph
\Ensure Causal graph is DAG, all nodes have properties
\While{true}
\If{operator is modifying causal graph}
    \State \textbf{Wait}
\Else
\State \textbf{Execute} with guidance from causal graph
        \State \textbf{Check} execution status
        \If{status is SUCCESS}
            \State \textbf{Continue}
    \Else
            \State \textbf{Notify} Operator
            \State \textbf{Ask} Continue?
        \EndIf
    \EndIf
\EndWhile
\end{algorithmic}
\end{algorithm}
\vspace{-2mm}

\subsection{Causal Graph Intervention}
For a pick-and-place task, we use our constructed causal graph to evaluate
whether the relationships between the defined variables are valid, which in
turn can be inferred by the robot's performance. For instance, if the robot
consistently fails to pick up a cordless drill but it has no issues with other
objects, then we can be relatively certain that several known properties or
relationships (e.g., weight, texture, etc.) may be detrimentally affecting the
robot. With this human-in-the-loop approach, we can quickly devise an
intervention procedure to decide if certain variables are causing degraded
performance. For example, we can intervene by setting the object's weight to a
specific value and then benchmark the robot's performance under the assumption
that the robot did not account for the object's weight properly. 

If the robot's performance is influenced by the weight of the object, then we
can make adjustments to the robot's behavior to ensure optimal performance
(e.g., adjusting the pick points, etc.). Alternatively, should the intervention
be inconclusive, then we can intervene by fixing other object properties to a
specific value and then benchmark the robot's performance. However, further
investigation may be required to understand the exact reason. For example, the
gripper might not be suited for the cordless drill's shape, or the perception
system may not be identifying the drill's correct pick points. Since our
framework has the capability to adjust causal graph structure on the fly, we
can quickly update the graph to reflect the new structure thus allowing the
robot to make more informed decisions.

\subsection{Discussion}
Our framework provides advantages over traditional methods for constructing
causal graphs. The AR interface allows an operator to interact with the robot
and the environment directly and more intuitively, making it easier for humans
to understand the system's dynamics. It also presents a visual understanding of
causal relationships, which allows operators to make informed decisions on how
to optimize the system. For example, the operator can visualize in real time
where interventions might be needed, or where potential failures could occur. 

Constructing causal graphs this way introduces a level of modularity and
scalability. For instance, causal elements can be easily adjusted to fit the
needs of the system or the entire graph can be replaced by constructing a
replacement from scratch. Yet, this methodology is not without its limitations.
A criticism of causal graphs is that they can be an oversimplification of the
environment and interplay between variables. Real-world robotic operations may
involve numerous subtle interactions, not all of which can be accurately
captured in a graph. 

\section{Conclusion}
\label{sec:conclusion}
This paper provided preliminary results on a human-centered approach for
automating the construction of causal graphical models. Our framework combines
the strengths of VAMR and simulation technologies to address the dilemma of
bootstrapping the creation of a causal graph. Furthermore, we highlighted the
ability to visualize, intervene, and update causal graphs on a physical robot
for a real-world pick-and-place task. In the future we will work on addressing
the challenges of oversimplification and iteratively updating causal graphs.

\bibliographystyle{ACM-Reference-Format}
\bibliography{a_human-centered_approach_for_bootstrapping_causal_graph_creation}

\end{document}